\documentclass[letterpaper, 10 pt, conference]{ieeeconf}  % Comment this line out if you need a4paper

\IEEEoverridecommandlockouts                              % This command is only needed if 
\usepackage{xcolor}
\usepackage{hyperref}
\usepackage{subcaption}   % For subfigures (for multiple plots in one figure)
\usepackage{tabularx}
\usepackage{multirow}
\usepackage{multicol}
\usepackage{ctable}
\usepackage{dcolumn}
\usepackage{balance}
\usepackage[T1]{fontenc}
\usepackage{graphicx}
\makeatletter
\renewcommand{\fnum@figure}{Figure \thefigure}
\makeatother

\title{\LARGE \bf
AI or Human? Understanding Perceptions of Embodied\\ Robots with LLMs

}

\author{Lavinia Hriscu$^{1,2}$ \and  Alberto Sanfeliu$^{1,2}$ \and Anaís Garrell $^{1,2}$       % <-this % stops a space
\thanks{This work has been supported by JST Moonshot R \& D (JPMJMS2011-85) and the European project TORNADO with grant number HORIZON-CL4-2024-DIGITAL-EMERGING-01-101189557.
}
\thanks{$^{1}$Institut de Robòtica i Informàtica Industrial (CSIC-UPC). Llorens Artigas 4-6, 08028, Barcelona, Spain.
        {\tt\small {lavinia.beatrice.hriscu,alberto.sanfeliu, anais.garrell}@upc.edu}}%
        \thanks{$^{2}$ Universitat Politècnica de Catalunya (UPC), Jordi Girona 31,         Barcelona, 08034, Spain.   
}
\thanks{This paper has been accepted by the 2025 34th IEEE International Conference on Robot and Human Interactive Communication (ROMAN) and the IEEE copyright notice.}
}

\usepackage{graphicx}
\usepackage{amsmath}

\begin{document}

\maketitle
\thispagestyle{empty}
\pagestyle{empty}

%%%%%%%%%%%%%%%%%%%%%%%%%%%%%%%%%%%%%%%%%%%%%%%%%%%%%%%%%%%%%%%%%%%%%%%%%%%%%%%%
\begin{abstract}

The pursuit of artificial intelligence has long been associated to the the challenge of effectively measuring intelligence. Even if the Turing Test was introduced as a means of assessing a system's intelligence, its relevance and application within the field of human-robot interaction remain largely underexplored. This study investigates the perception of intelligence in embodied robots by performing a Turing Test within a robotic platform. A total of 34 participants were tasked with distinguishing between AI- and human-operated robots while engaging in two interactive tasks: an information retrieval and a package handover. These tasks assessed the robot’s perception and navigation abilities under both static and dynamic conditions. Results indicate that participants were unable to reliably differentiate between AI- and human-controlled robots beyond chance levels. Furthermore, analysis of participant responses reveals key factors influencing the perception of artificial versus human intelligence in embodied robotic systems. These findings provide insights into the design of future interactive robots and contribute to the ongoing discourse on intelligence assessment in AI-driven systems.

\end{abstract}
%DESTINATION:
%We combine incremental learning methods with dynamic hand gesture recognition and introduce a novel data processing method that allows continuous learning of new dynamic gestures in real-time.
%ROADMAP:
%- Compare our method against SOTA methods on two datasets for Isolated DHGR
%- Demonstrate on two datsets that our architecture is capable of maintain knwoleadge when new classes are learnt (IL).
%- Demonstrate that our data processing pipeline can be used in real-time scenarios and improves existing results, we also integrate our solution in a real robot.

%%%%%%%%%%%%%%%%%%%%%%%%%%%%%%%%%%%%%%%%%%%%%%%%%%%%%%%%%%%%%%%%%%%%%%%%%%%%%%%%
\section{Introduction} \label{sec:introduction}

The theory of multiple intelligences identifies eight distinct cognitive domains \cite{intelligence}, with linguistic intelligence playing a fundamental role in human communication. Recent advancements in artificial intelligence (AI) have led to the development of Large Language Models (LLMs), trained on vast amounts of textual data to generate contextually relevant, human-like responses \cite{zhao2023}. The integration of LLMs into robotics has significantly expanded the capabilities of intelligent systems \cite{zeng2023fanlong}, facilitating applications such as trajectory planning \cite{bucker2023latte}, robot control \cite{llmcontrol}, emotional interaction \cite{mishra2023real}, social cooperation \cite{zhang2023large}, and task planning \cite{singh2023progprompt}.  

However, fundamental questions remain regarding the perceived intelligence of LLM-driven robots and their ability to exhibit human-like behavior in real-world scenarios. Research on Human-Robot Interaction (HRI) suggests that embodiment—the physical presence of a system—enhances natural communication by integrating verbal, non-verbal, and behavioral cues \cite{turingembod}. Therefore, LLM-equipped robots are prime candidates for showing multiple forms of intelligence.

This discussion ties into a broader debate on machine intelligence. Alan Turing’s seminal 1950 work introduced the Turing Test as a measure of artificial intelligence, assessing a system’s ability to manifest human-like conversational behavior \cite{turing}. This test has since sparked ongoing discourse regarding its validity and the extent to which machines can genuinely replicate human cognition \cite{french}. In the context of LLM-driven robots, the Turing Test remains a relevant framework for evaluating whether humans can reliably distinguish between artificial and human intelligence.

\begin{figure}[t]
 \centering
  \subfloat[]{
%\hspace{-0.3cm}
   \label{person_robot}
    \includegraphics[width=0.45\columnwidth]{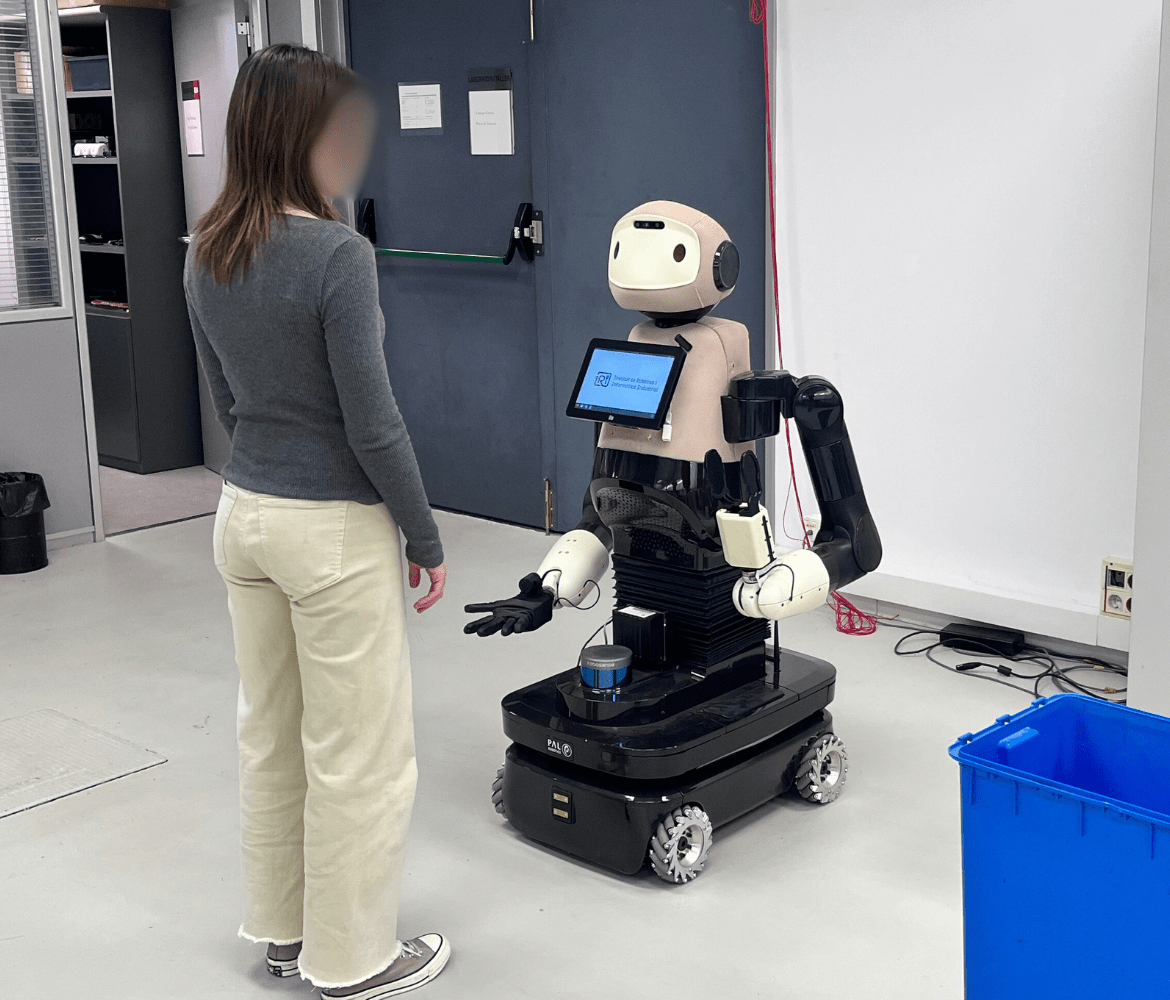}}
  \subfloat[]{
    %\hspace{-0.2cm}
   \label{person_package}
    \includegraphics[width=0.45\columnwidth]{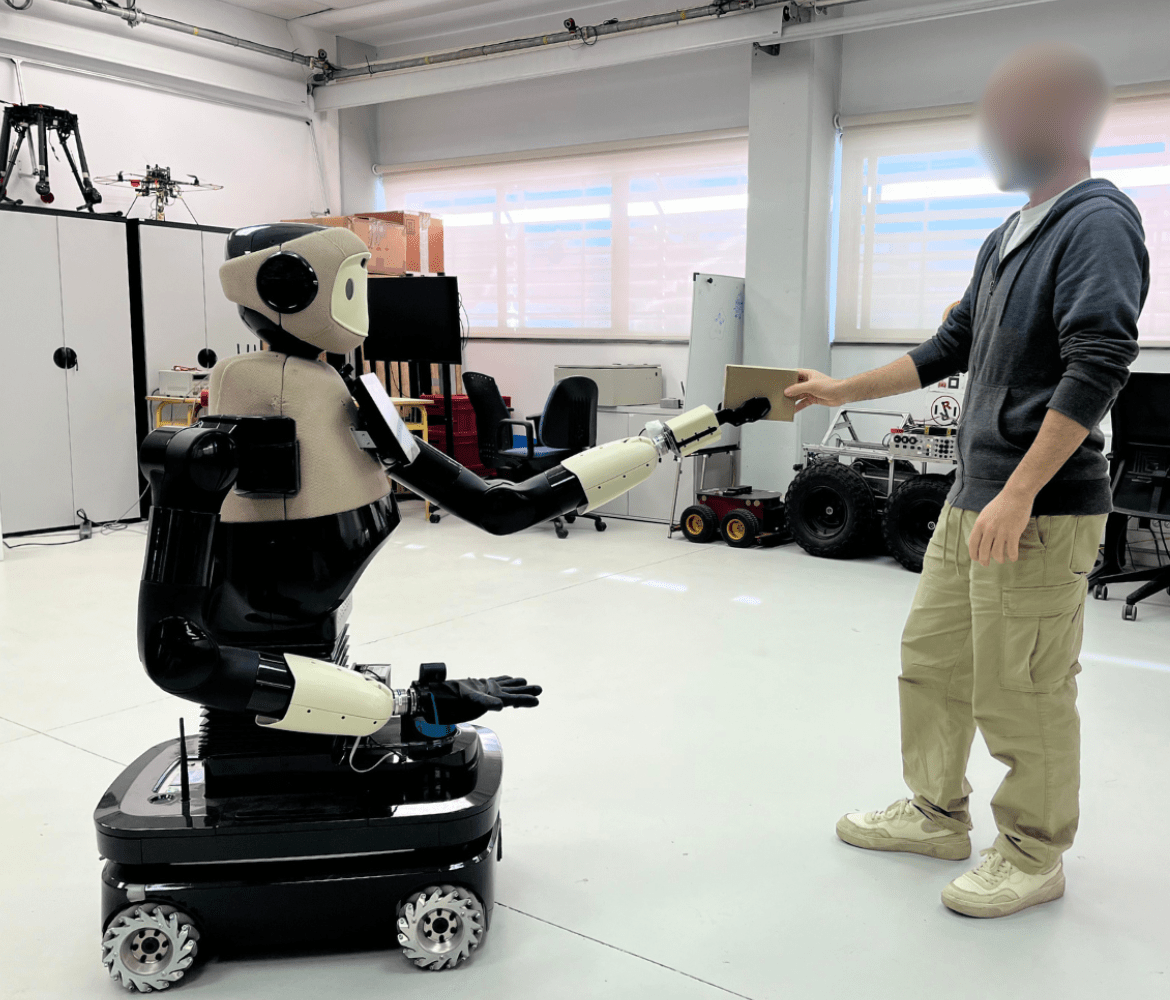}}\\ \vspace{0.1cm}
  \subfloat[]{
 % \hspace{-0.35cm}
   \label{robot_package}
    \includegraphics[width=0.45\columnwidth]{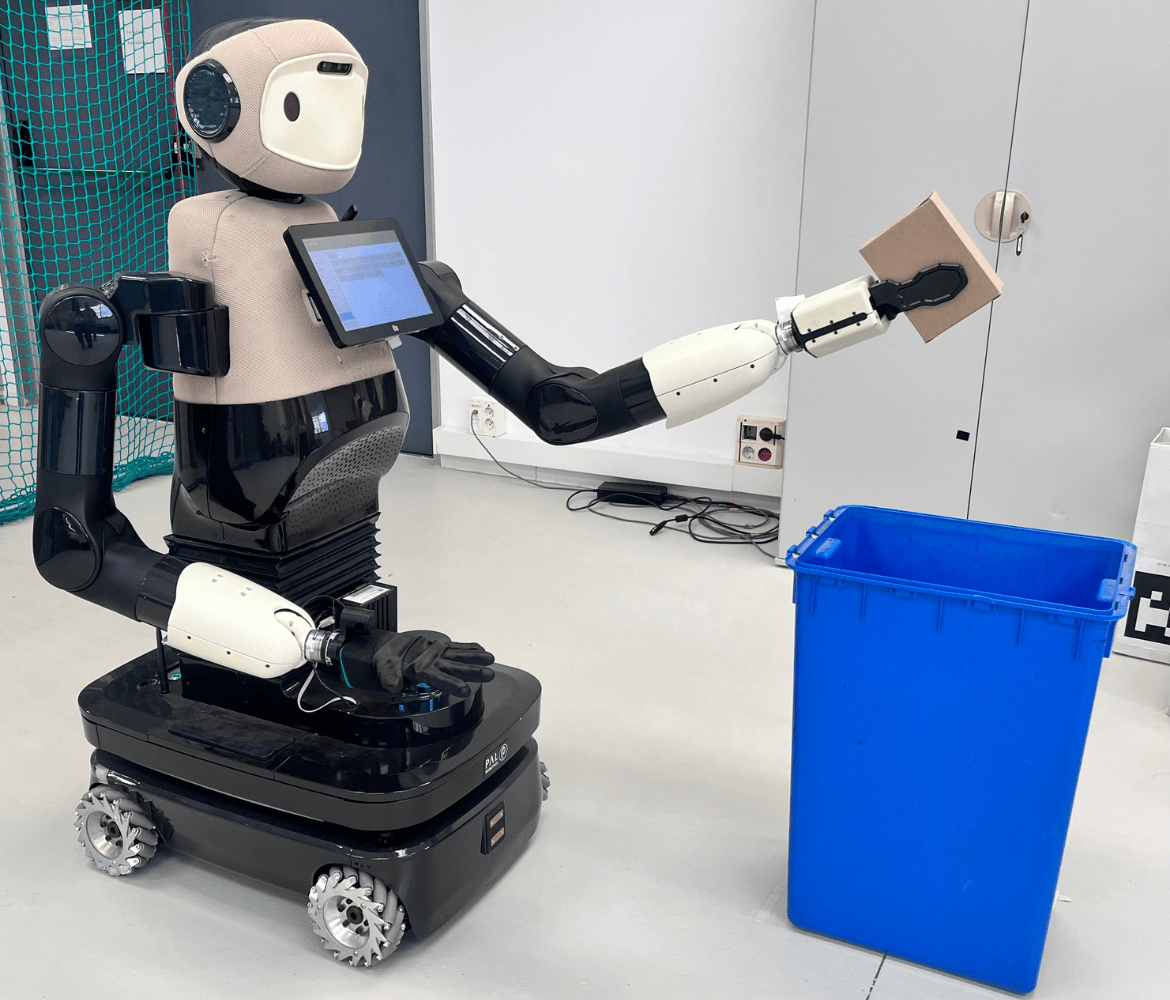}}
    \subfloat[]{
 % \hspace{-0.35cm}
   % \hspace{-0.2cm}
   \label{operator_interface}
    \includegraphics[width=0.45\columnwidth]{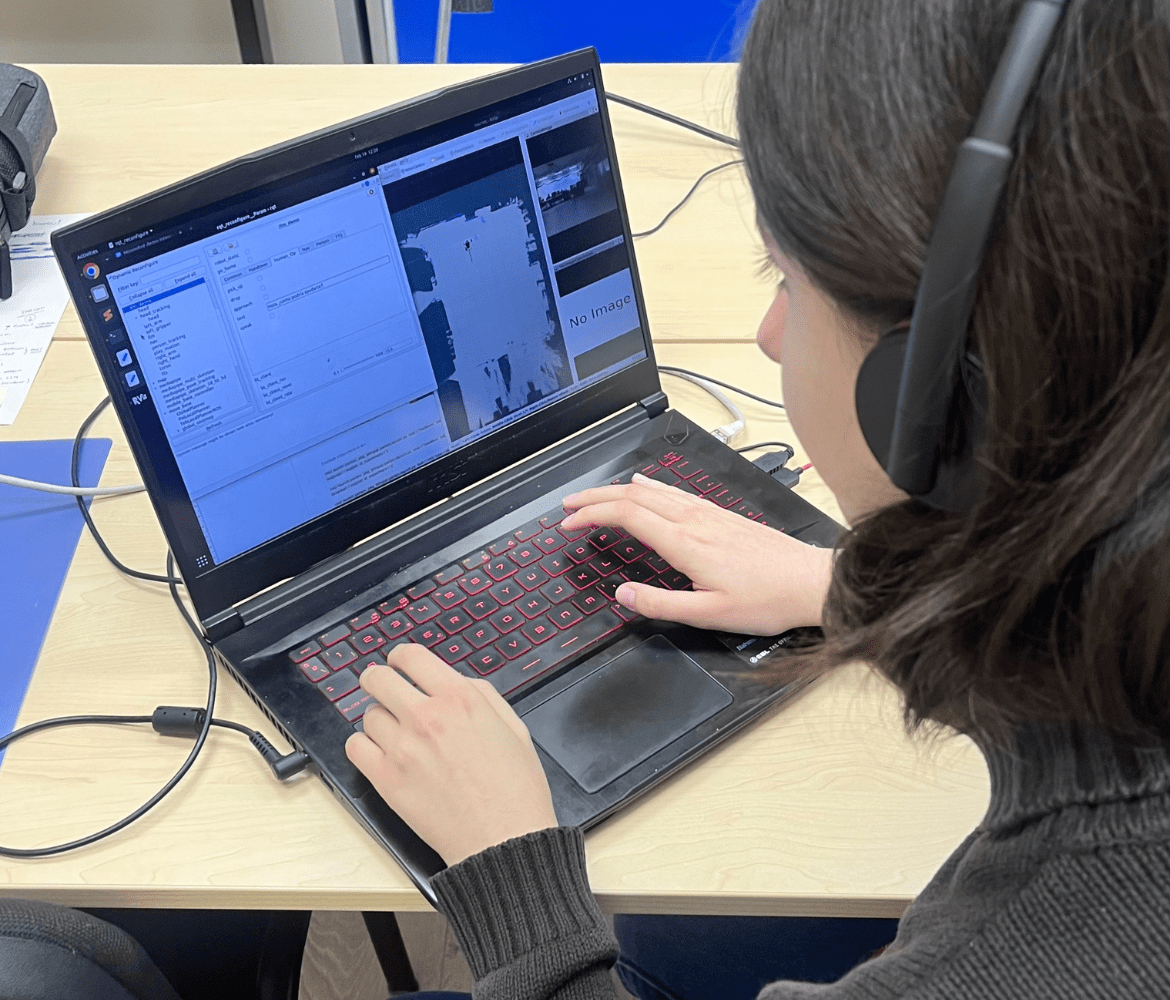}}
 \caption{(a) Participant engaging with the robot in static mode for the information assistance task at the initial position. (b) Participant interacting with the robot in dynamic mode during the package delivery task after the robot approached him. (c) Robot delivering the package in dynamic mode. (d) Operator controlling the robot through the interface.}
 \label{fig:real_scenario}
\end{figure}

Motivated to explore how and why humans perceive AI-driven systems differently from human-operated ones, we designed an embodied Turing Test using a robot controlled by either a human or an AI operator, powered by an LLM at its core. Our study addresses two key research questions: (1) Are humans better than chance at correctly detecting the operator type for an embodied robot? (2) Is one type of operator misidentified more frequently than the other?

To explore these questions, 34 participants interacted with a robot across two tasks: information assistance and object delivery, the latter requiring robotic grasping. The robot operated in two modes: static, where only its arm moved, and dynamic, which included person detection and navigation. Each participant engaged with the robot in both modes and for both tasks. Participants completed a questionnaire covering demographic information, a Turing Test assessment, and their reasoning behind their responses. In this paper, we present our experimental findings and contributions:

\begin{enumerate}
    \item Assess human ability to distinguish between AI and human operators and identify factors influencing this perception.
    \item Showcase the reasons behind participants responses, guiding the development of more human-like and trustworthy AI attitudes.
    \item Compare human- and AI-controlled robots behaviors in real-world task scenarios.
    
\end{enumerate}

The remainder of this paper is organized as follows: Section II reviews related work, Section III explains the method, Section IV reveals the results, Section V presents a discussion of the findings, and Section VI provides the conclusions of this study.

\section{Related Work} 
 
This section examines related research on the Turing Test within the realm of intelligent systems, with a particular focus on its applicability to LLMs. Subsequently, we review teleoperated robots, highlighting their potential for comparing human-controlled and AI-driven interactions.

\subsection{Turing Test applied to intelligent systems}

Over the years, the Turing Test has evolved into multiple approaches, each emphasizing different aspects of intelligence. Several studies have explored methods for evaluating intelligence in multimodal settings \cite{turingiathlon, turingforbus}, with some focusing on visual question answering \cite{turingmultimodal, turingvisual}, while others have investigated emotional traits \cite{turingsocial}.  

Although the original formulation of the test centered solely on text-based communication, contemporary interpretations have examined the role of embodiment in assessing artificial intelligence. Embodied systems with intelligent capabilities have emerged as a means to evaluate intelligence more comprehensively, considering four key aspects: language, perception, reasoning, and action \cite{turingortiz}. Research indicates that embodiment enhances user engagement, empathy, trust, and perceived intelligence of robotic agents \cite{turingembod1, turingembod2, turingembod3, turingembod4}. Notably, \cite{turingnonverbal} explores how incorporating human-like variability into a humanoid robot could enable it to pass a nonverbal Turing Test.

In our work, we conducted an embodied verbal Turing Test using a robot capable of spatial navigation, object and human perception, verbal interaction, and object manipulation. This approach allows us to assess not only communication aspects but also task-oriented capabilities that require movement, such as approaching people and collecting packages.

\subsection{Turing Test applied to LLMs}

The advanced capabilities of LLMs position them as strong candidates for serving as intelligent communication components in embodied systems \cite{wang2024large}. Previous studies have investigated Turing Tests involving LLM-driven conversations on online platforms \cite{turingllms, turinggpt4}. Expanding on these findings, \cite{turingpass} demonstrates that distinguishing a GPT-4 model from a human becomes increasingly difficult.  

Along similar lines, researchers have also examined the detection of artificially generated content \cite{turingbench, sciencellm, surveyllm}. In particular, \cite{inverted} examines the effect of removing real-time interaction with the model, instead using pre-recorded conversations for evaluation.

Our study examines how LLMs should behave to be perceived as human-like as possible in real-time interactions, aiming to pass the Turing Test. In our case, these models not only generate responses in a dialogue but also trigger actions in a robot, enabling more dynamic and real interactions.

\subsection{Teleoperated robots}

In recent years, teleoperation applications for robots have expanded across various fields, including surgery \cite{teleopsurgery, teleopsurgery1}, high-risk human-contact environments \cite{teleopcovid, teleopcovid1}, and social robotics \cite{teleopcoach, teleopass}. In the context of an embodied Turing Test, teleoperation provides a valuable baseline for comparing AI-driven robotic behaviors with those controlled by human operators. 

To the best of our knowledge, no prior research has explored applying the Turing Test to robots powered by LLMs or teleoperated by humans in conversational tasks. To bridge this gap, we developed an interface for controlling a robot in real-time interactions, enabling a comparative analysis of LLM-driven and human-operated behaviors.

\section{Method}

In this study, we examine whether users can distinguish between a robot controlled by a human operator and one driven by artificial intelligence, while also exploring the key factors that shape this perception. The experiments were conducted with the approval of the Ethics Committee of the Universitat Politècnica de Catalunya (UPC), adhering to all relevant ethical regulations and guidelines (ID: 2024.020).

\begin{figure*}[t]
 \centering
  \subfloat[]{
   \label{reasonai}
     %\hspace{-1cm}
    \includegraphics[width=0.49\textwidth]{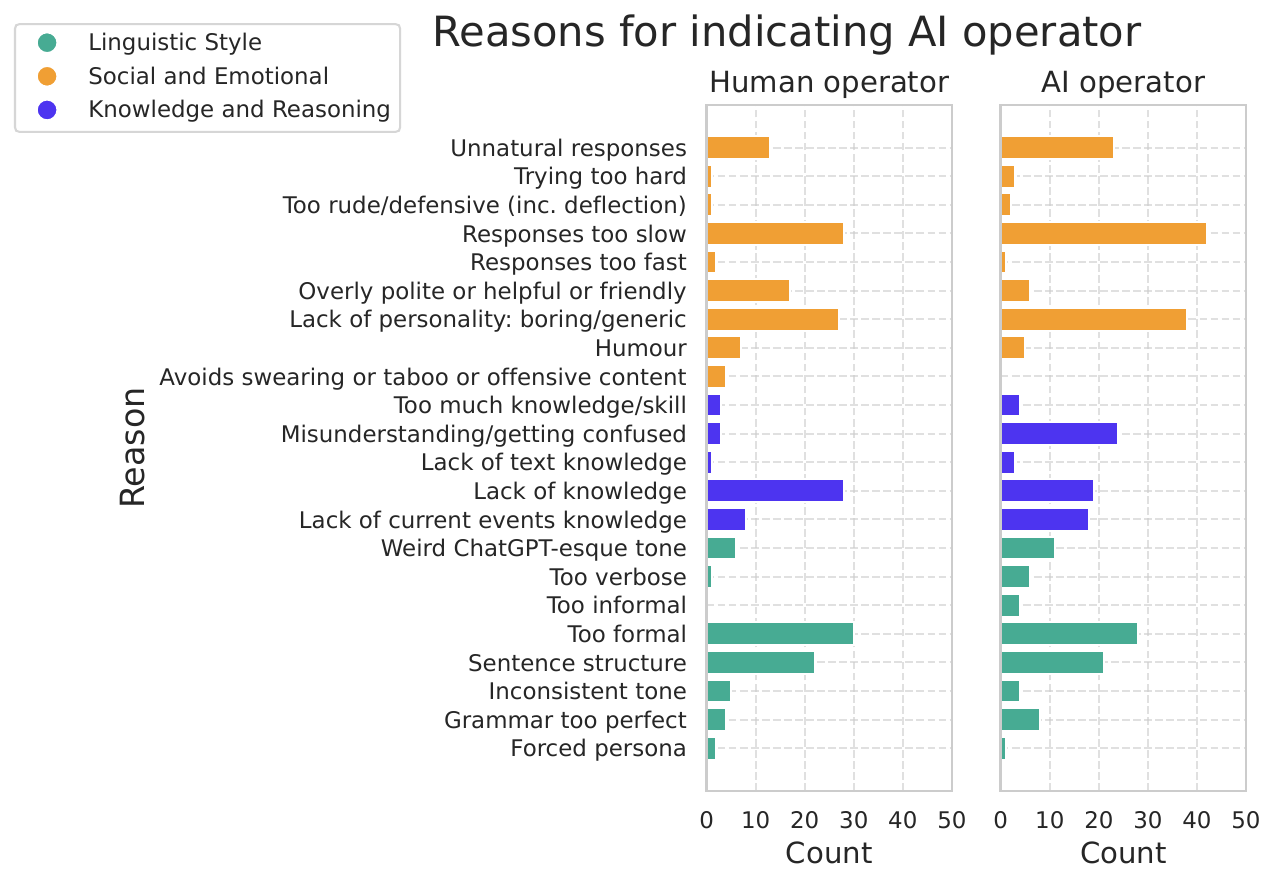}}
  \subfloat[]{
  \hspace{-0.4cm}
   \label{reasonhum}
    \includegraphics[width=0.455\textwidth]{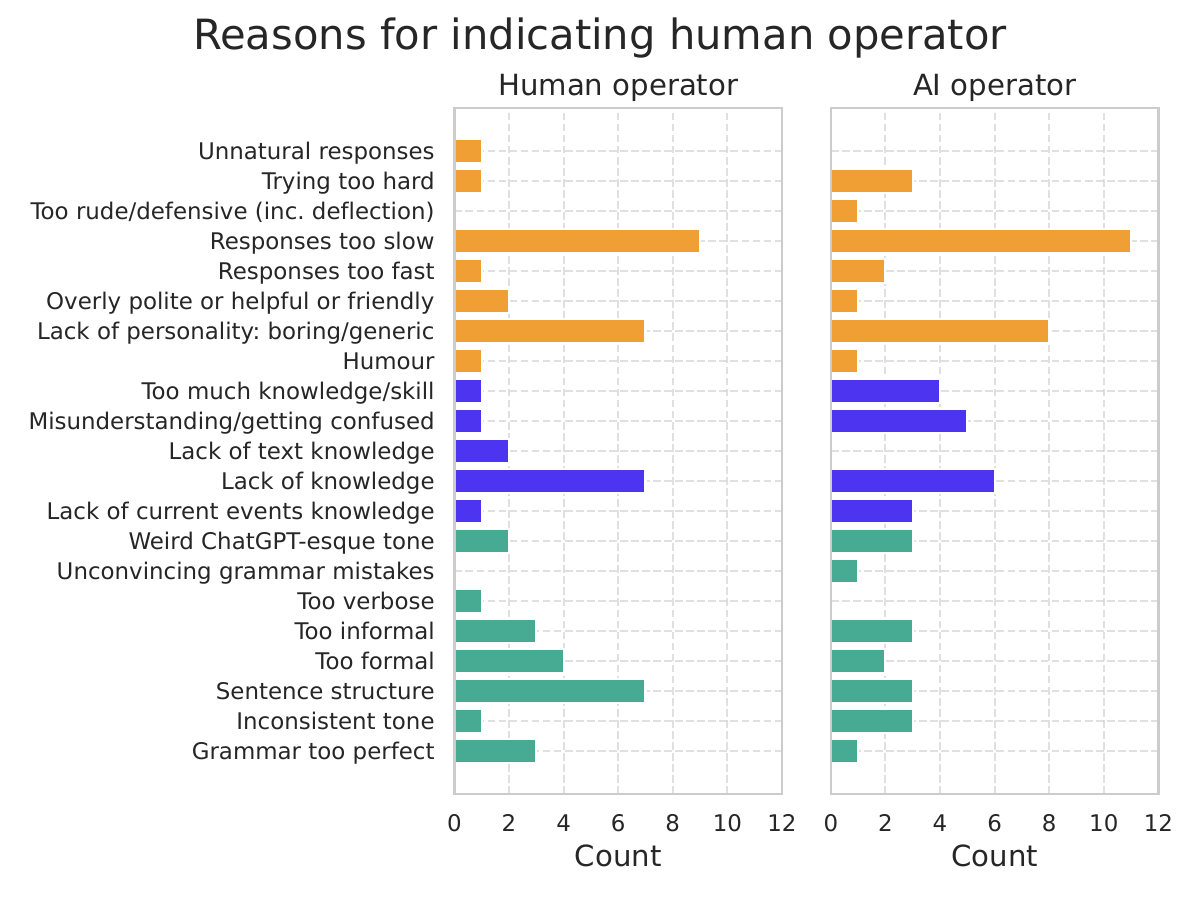}}
 \caption{Bar charts comparing the reasons participants attributed to perceiving an AI operator (a) or a human operator (b) as the answer to the Turing Test. For both answers, the plots are segmented by the actual type of operator (human or AI). Reasons are grouped by category: Linguistic Style, Social and Emotional, and Knowledge and Reasoning.}
 \label{fig:graph_reasons}
\end{figure*}

\subsection{Embodiment design}

The experiments were conducted using IVO \cite{laplaza}, a bimanual robot capable of grasping objects, navigating its environment, and interacting through vision, hearing, and speech. All experiments took place at the Mobile Robotics Laboratory of the Institut de Robòtica i Informàtica Industrial CSIC-UPC in Barcelona, Spain. 

To enable IVO's listening, response generation and speech capabilities, the following models and configurations were used to create nodes in ROS 2 Hawksbill~\cite{rosdoc}:

\begin{itemize}
    \item Speech-to-Text (STT): Vosk models for English (vosk-model-en-us-0.22), Spanish (vosk-model-es-0.42), and Catalan (vosk-model-small-ca-0.4) \cite{STT}.
    \item Large Language Model (LLM): OpenAI GPT-4o-mini \cite{chatgpt} with a temperature setting of 1.
    \item Text-to-Speech (TTS): gTTS (Google Text-to-Speech) \cite{TTS}.
\end{itemize}

The robot utilized its built-in navigation module for autonomous movement. For person detection and tracking, it employed the \textit{spencer\_people\_tracking} package from the Spencer project \cite{spencer}. Both components ran within the ROS framework, with a ROS Bridge facilitating seamless integration with ROS 2. 

\subsection{Experiment design}

The experiment involved two distinct movement modalities for the robot. In dynamic mode, the robot autonomously detected and approached the participant, continuously tracking their head movements throughout the interaction. In static mode, the robot remained stationary, with the only action being the extension of its arm to receive objects.  

Each participant completed two tasks, both performed in each movement mode, making task type and movement mode within-subject factors. Participants engaged in single-turn interactions and had the option to choose from three languages—Catalan, English, or Spanish—based on their local language familiarity. The task order was randomized, with both modes of the same task performed consecutively.

\subsubsection{Task 1: Information assistance}

In this experiment, the robot serves as an information assistant. Both the human operator and the LLM had access to a document with fictional details about a university building. The LLM used Retrieval Augmented Generation (RAG) \cite{lewis2020retrieval} with the Language-agnostic BERT Sentence Embedding (LaBSE) model \cite{labse} to match queries in three languages. The operators were instructed to respond solely on the basis of the document's content, avoiding unrelated information. Participants were informed about the university context and encouraged to ask typical concierge questions. They could interact and ask questions as desired, as seen in \autoref{person_robot}.

\subsubsection{Task 2: Package delivery}

In this task, the participant received a small package and interacted with the robot to decide whether to hand it over. No specific instructions were provided, allowing the participant to manage the interaction independently. If the participant agreed to deliver the package, the robot extended its arm to receive it, as shown in \autoref{person_package}. In static mode, the robot held the package in place, while in dynamic mode, it moved to a drop-off point, deposited the package, and returned to initial position, see \autoref{robot_package}.

\subsection{Operators}

The experiment focused on the content of the responses, with both operators using the same voice and sending their replies as text to the Text-to-Speech (TTS) system of the robot. Movement modules, integrated into a behavior tree, remained unchanged and were triggered by either the human operator through an interface, or the LLM with action flags.

\subsubsection{Human Operators}

A pilot study with nine operators (eight male, one female) was conducted to analyze human operator behavior and response times to refine instructions. Each operator answered six randomly selected questions—three written and three verbal. Verbal response mode was discarded due to transcription errors from STT conversion prior to the robot's TTS.

Two additional operators (one male, one female) participated in 5 and 12 experiments, respectively, controlling the robot remotely via Rviz \cite{rviz}, without physical contact with participants, as shown in \autoref{operator_interface}. Operators were briefed on their role in controlling IVO and were given task instructions. For information assistance, they were provided with the same document as the LLM, along with Catalan and Spanish translations, and instructed to give accurate responses. For package delivery, operators were asked to establish trust with participants and complete the delivery successfully.

\subsubsection{AI Operator}

The LLM aimed to closely mimic human operator behavior, with human responses from the pilot study used as a reference to create the prompts. Each task had a distinct prompt, but all shared common elements: the robot’s name (IVO), conversation history, user input, interaction language, and university concierge role. Both prompts were in English, but instructions and interaction styles varied.

\begin{itemize}
    \item Information assistance: IVO participates in a Turing Test, responding strictly based on context provided by RAG. It ensures linguistic consistency, contextual accuracy, and concise responses, without revealing its AI nature or inferring information.
    \item Package delivery:  IVO facilitates efficient package handovers at the university entrance, using minimal and informal conversation. It avoids small talk and personal topics, and it only mentions the package if the human brings it up first.
\end{itemize}

The pilot study revealed that mimicking both linguistic behavior and response times was crucial. We compared the LLM's response times to those of human operators. To replicate typing speed, we used a method similar to that in \cite{turingpass}, where \( 0.3 \) represents seconds per character, \( n\_char \) is the number of characters of the input, and the term \( 1 \)  ensures a minimum delay in seconds:

\vspace{-0.3cm}
\begin{equation}
    delay = 1 + \mathcal{N}(0.3,0.03)\times n\_char
\end{equation}

\subsection{Participants}

We recruited 34 participants (24 male, 10 female), aged 18 to 64 years (mean = $33.23$, std = $10.88$). Participants were informed that they would participate in a Turing Test, in which they would interact with either an AI or a human operator. Each participant interacted with only one type of operator across both tasks and movement modalities, with 17 assigned to each group to balance the operator type as a between-subjects factor.

Participants were instructed to base their judgments solely on the interaction and avoid external assumptions. They selected their preferred language of interaction: 14 chose Catalan, 10 chose Spanish, and 10 chose English. The operator type was balanced to ensure 7 interactions in Catalan, 5 in English, and 5 in Spanish, for both operators. All participants provided informed consent via an institution-approved form.

% Explicar quantes noies, nois per cada idioma, tipus d'operador?

\subsection{Questionnaire design}

Before the experiments, participants answered demographic questions. After each interaction, they completed a Turing Test-based questionnaire adapted from \cite{turingpass}, indicating whether they interacted with an AI or a human operator, rating their confidence (1-10), and selecting reasons for their decision. After all four interactions, participants could provide feedback.

\section{Results}

In this section, we begin by addressing the two research questions outlined in \autoref{sec:introduction}. Next, we analyze test accuracy and explore the relationship between operator type, demographic factors, and conversational data. Finally, we conclude with a brief discussion on hallucinations in AI interactions.

Considering the four experiments conducted by each participant, we obtained a total of 136 samples. Analyzing participants' responses to the operator type question, we found 71 correct identifications (52.21\%) while 65 were incorrect (47.79\%). Breaking down the misidentifications by operator type, participants incorrectly identified a human operator as an AI in 50 instances (36.76\%) and mistook an AI operator as a human in 15 cases (11.03\%).

\begin{figure}[t]
      \centering
      \includegraphics[width=\columnwidth]{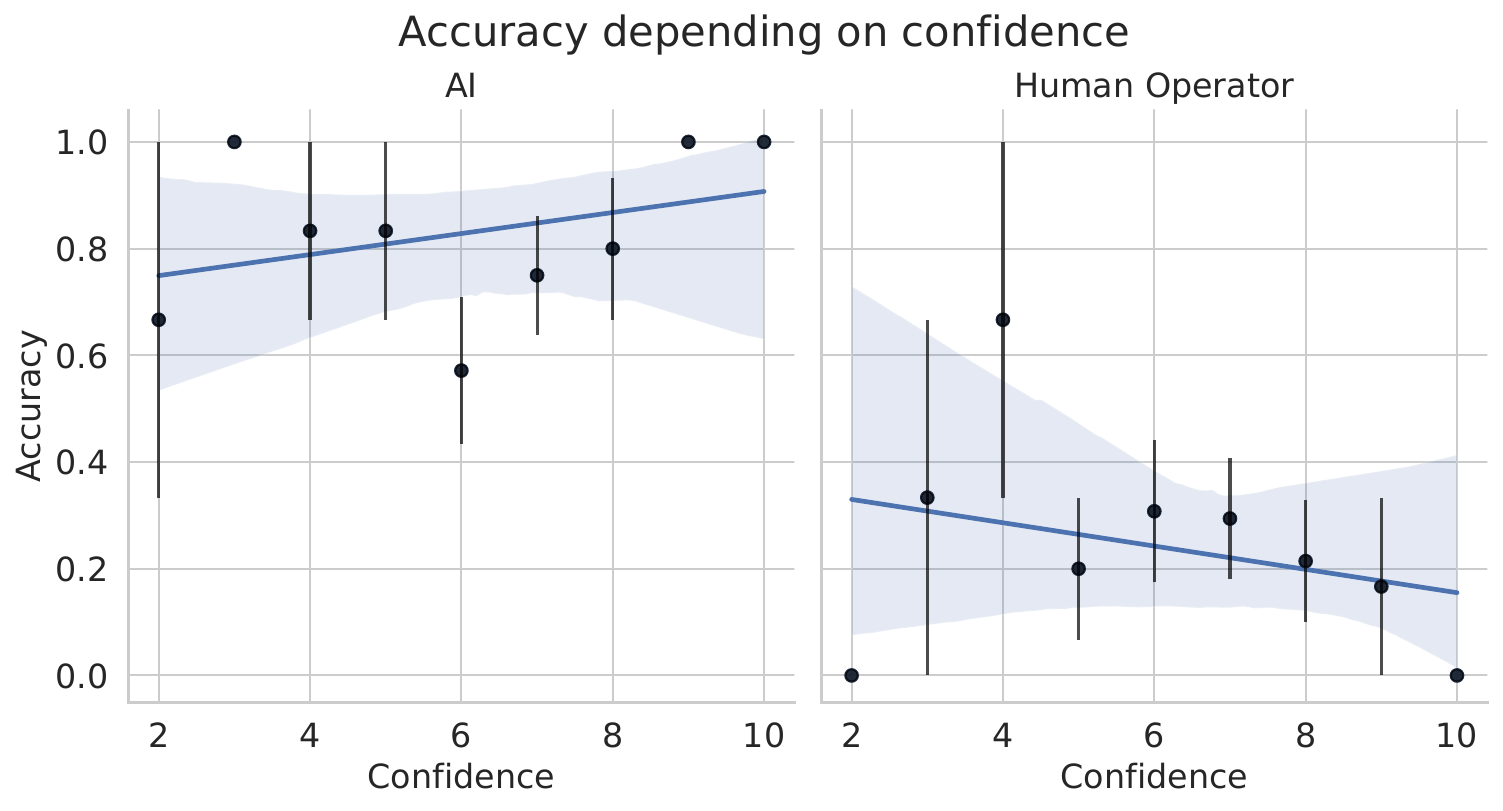}
      \caption{Scatter plots of the confidence levels in relation to participants' accuracy, segmented by real operator type. The shaded area represents the 95\% confidence interval.}
      \label{fig:graph_confidence}
   \end{figure}
   
Before analyzing each section, we note that no differences were found in accuracy, nor in misidentification rates by actual operator type across tasks and modalities, so no further results on this matter will be presented.

\subsection{Hypotheses}

For the first research question, we hypothesized that humans would not perform better than chance in detecting the operator type for an embodied robot. A binomial test with a success probability of $0.5$ showed no significant deviation from 50\% ($p=0.334$), indicating participants couldn't distinguish between human and AI operators above chance.

The second research question addresses the imbalance between the rate errors segmented by actual operator type. We hypothesized no difference in misidentification rates between operator types. A Chi-square test at a $5\%$ significance level showed a significant difference in misidentification rates ($\chi^2=34.07$, $p=5.33\times10^{-9}$), leading us to reject the null hypothesis.

\begin{figure*}[t]
      \centering
      \includegraphics[width=\textwidth]{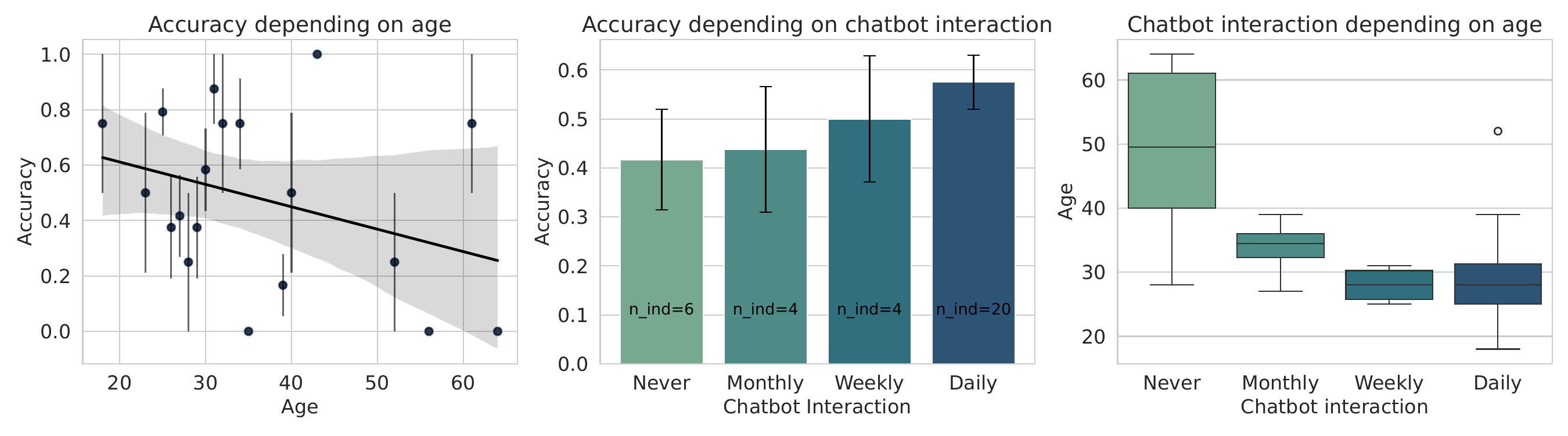}
      \caption{Relationships between age, level of chatbot interaction, and accuracy in the Turing Test. The left plot shows a regression line indicating the correlation between accuracy and age. The center plot presents how accuracy varies based on chatbot interaction frequency, with $n\_ind$ representing the number of participants that selected each frequency option. The right plot illustrates the distribution of age across different levels of chatbot interaction.}
      \label{fig:triplot_acc_age_chatbot}
   \end{figure*}
   
\subsection{Confidence and reasons}

Participants identified reasons for determining the operator type in each interaction. In \autoref{fig:graph_reasons} we compare the reasons selected by operator type. We focus on "Response too slow" due to the added delay mimicking human response times. This reason was selected 37 times for the human operator and 53 times for the AI operator. Of 34 participants, 17 selected it for the AI operator and 13 for the human operator, with only 4 participants never selecting it. Additional feedback also indicates that most participants wanted the robot to offer suggestions or alternatives when it didn’t know the answer.

The relationship between confidence and accuracy is shown in \autoref{fig:graph_confidence}. For the AI operator, accuracy increases with higher confidence, with relatively concentrated data points and some variability. For the human operator, accuracy decreases as confidence rises, with more dispersed data points and a wider confidence interval, indicating greater variability. Many participants mentioned that increased interaction with the robot made them feel more comfortable and better understand its capabilities.

\subsection{Demographic data}

We visualized the relationship between participants' responses and factors such as gender, preferred language, education level, knowledge of robotics and LLMs, and number of prior interactions. No patterns emerged to suggest further analysis. However, we conducted a more detailed analysis of age and level of chatbot interaction, as shown in \autoref{fig:triplot_acc_age_chatbot}. Accuracy tends to decrease with age, while younger participants engage with chatbots more frequently. A correlation also appears between higher familiarity with chatbots and greater accuracy. However, it is important to note that most participants interacted with chatbots daily.

\subsection{Conversational data}

\begin{figure}[h!]
      \centering
      \includegraphics[width=0.49\textwidth]{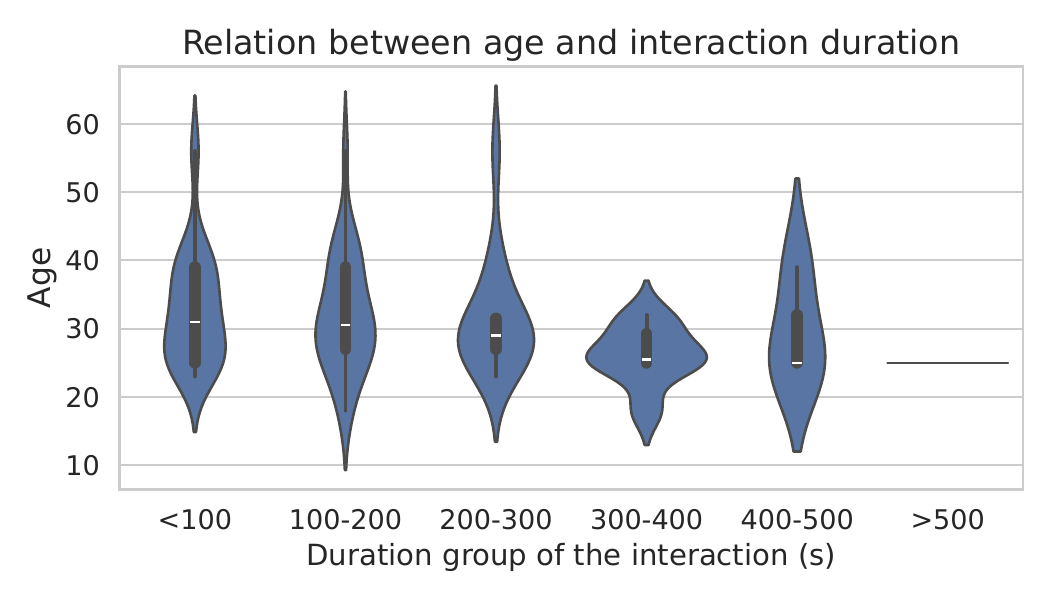}
      \caption{Violin plots with the age of the participants and the duration group of their interactions with the robot.}
      \label{fig:graph_age_duration}
\end{figure}

Quantitative data was collected for each interaction. We calculated the total duration time and categorized it into different duration groups as presented in \autoref{fig:graph_age_duration}.

In \autoref{tab:times}, we present the average duration and number of single-turn interactions (STI), segmented by operator, task, and modality. Human operators had shorter durations but more single-turn interactions, especially in the information task, which took longer than the delivery task for both operators.

\begin{table}[h]
\center
\begin{tabular}{|c|c|c|c|c|}
 \cline{1-5}
 & \multicolumn{2}{c|}{AI operator} & \multicolumn{2}{c|}{Human operator} \\ \cline{2-5}
 Interaction & Time (s) & STI & Time (s) & STI \\  \cline{1-5}
 HD & 141.23 & 5 & 112.29 & 6.78 \\ \cline{1-5}
 HS & 108 & 4.29 & 98.46 & 6 \\  \cline{1-5}
 ID & 227.99 & 5.69 & 221.75 & 10 \\ \cline{1-5}
 IS & 222.53 & 5.69 & 262.67 & 11.83 \\ \cline{1-5}
\end{tabular}
 \caption{Average number of single-turn interactions, and average total duration time for the interactions with both types of operator, segmented by task and modality: handover dynamic (HD), handover static (HS), information dynamic (ID), information static (IS).}
 \label{tab:times}
\end{table}

\subsection{Hallucinations}

We define two types of hallucinations based on user queries. Known hallucinations occur when questions relate to content provided to the LLM. If the LLM gives an incorrect response or states it lacks the information, it is classified as a known hallucination. No known hallucinations were observed during the experiments.

Unknown hallucinations occur when questions concern content not provided to the LLM. If the LLM responds anything instead of indicating it lacks the information, it is classified as an unknown hallucination. There were only 8 such instances out of 64 interactions. Interestingly, participants correctly identified the operator type in 6 of the 8 cases when a hallucination occurred.

\section{Discussion}

In this study, we explored humans' ability to distinguish between human and AI robot operators based on their verbal interactions with a robot. $34$ participants engaged in the experiment, which involved an information retrieval task and a package handover task, evaluating the robot’s perception and navigation abilities under both static and dynamic conditions. Our findings indicate that participants struggled to identify the operator type beyond chance level, with a pronounced tendency to misclassify human operators as AI. Moreover, task type and robot’s movement mode had no significant impact on identification accuracy. In this section, we explore how participant demographics influence accuracy, examine perceptions of each operator type, and analyze the key factors shaping these judgments.

\vspace{-1mm}

\subsection{Performance results and demographics influence}

Our analysis uncovers a complex relationship between perception, confidence, and accuracy when distinguishing between AI and human operators. Although in \autoref{fig:graph_reasons} participants seem to have multiple reasons for classifying an operator as AI or human, their overall accuracy remained at chance levels, suggesting they were unable to reliably identify the operator type.

The confidence graph in \autoref{fig:graph_confidence} suggests that participants had a clearer understanding of how AI communicates. However, when classifying human operators, increased confidence was associated with decreased accuracy. This could imply that participants held rigid or incorrect assumptions about human communication or that human operators did not fully behave as naturally as expected.

Furthermore, the results in \autoref{fig:triplot_acc_age_chatbot} reveal that familiarity with chatbot systems, particularly among younger participants, shaped their perceptions while rising slightly their accuracy. Although repeated interactions increased their comfort and understanding of chatbot capabilities, participants' ability to correctly identify the operator remained unchanged. Accuracy was consistent across interactions, suggesting that mere exposure was not enough to improve their ability to differentiate between AI-generated and human responses.

These findings emphasize a critical challenge: people may tend to rely on assumptions that don't always align with the actual differences between AI and human communication. This raises important questions for AI design: Should systems aim to mimic human interaction more closely, or should they maintain distinct characteristics to avoid confusion? Additionally, the results highlight the importance of becoming familiar with AI systems, as this may help people better differentiate AI-generated content, which is sometimes misleading, from content generated by humans.

\vspace{-1mm}

\subsection{Perception and real behavior of operators}

The pilot test showed that human operators delivered more direct and unrestricted responses, whereas the LLM, without an adjusted prompt, demonstrated a more rigid conversational style.

As illustrated in \autoref{fig:graph_reasons}, both response time and personality significantly shape user perception, regardless of the operator type. According to \autoref{tab:times}, human operators either provide shorter responses, reply more quickly, or both, allowing them to engage in more interactions within the same total time. Interestingly, response delay was strongly linked to perceiving AI, being cited 70 times compared to 20 for humans. Despite AI operators receiving more votes overall, the ratio remains similar, suggesting that, while AI responses were perceived as slower, the actual difference in response times between AI and human operators was not substantial.

Linguistic style also plays a crucial role in distinguishing AI from human operators. AI-generated responses are often expected to be overly formal and structured, while informal and natural dialogue is linked to human interaction. However, paradoxically, human operators in our study exhibited even stronger formal tendencies than the AI, challenging common assumptions about conversational tone.

An intriguing aspect of our findings is that human operators, who were instructed to follow specific guidelines, led participants to question if their behavior had been intentionally shaped to resemble AI. AI is often associated with responses that are unnatural, excessively polite, and impersonal. However, participant's feedback suggests that when human operators strictly follow the guidelines, their behavior at times resembles that of machines. This may partly explain the notable misidentification rate among operators and, in turn, raises a broader question in the ongoing AI development debate: Should AI be designed to replicate human emotions and expressiveness, or should it adopt a more neutral, detached communication style?

Furthermore, our findings emphasize the importance of adapting AI systems to new data and personalizing interactions based on user profiles. The tendency of AI to struggle with recent events, spatial awareness and to get confused in conversations was a major indicator of its artificial nature. Addressing these shortcomings requires techniques such as fine-tuning, RAG, multimodal models and AI agent frameworks, which enhance adaptability and contextual awareness—capabilities that are essential for AI to more closely approximate human-like interaction. In our study, RAG has proven to be efficient considering the low rate of hallucinations.

%\section{FUTURE WORK}
% Seria comentar que s'hauria de veure sense delay, afegint el temps de resposta com una variable que tingues efecte real en la resposta. I també que l'operador respongués per veu.
\vspace{-1mm}
\section{Conclusions}

This study examines humans' ability to distinguish between AI and human robot operators through conversations in information retrieval and package delivery tasks. The results show that participants were unable to identify the operator type above chance levels, with response time and linguistic style influencing their judgments. Accuracy was affected by age and chatbot interaction levels, but remained unaffected by task type, robot movement, and other demographic factors. Despite displaying more personality, human operators’ responses were more frequently misidentified as AI, rather than the reverse, highlighting the tendency to associate AI-generated content with formality and politeness. Retrieval-Augmented Generation  was effective in reducing errors, improving AI response quality. However, when AI produced hallucinations, these clearly revealed its artificial nature. The findings illustrate the subtle challenges humans face in distinguishing AI from humans in conversation, emphasizing the role of linguistic cues and user experience.

\addtolength{\textheight}{-2cm}   % This command serves to balance the column lengths
                                  % on the last page of the document manually. It shortens
                                  % the textheight of the last page by a suitable amount.
                                  % This command does not take effect until the next page
                                  % so it should come on the page before the last. Make
                                  % sure that you do not shorten the textheight too much.

%%%%%%%%%%%%%%%%%%%%%%%%%%%%%%%%%%%%%%%%%%%%%%%%%%%%%%%%%%%%%%%%%%%%%%%%%%%%%%%%

%%%%%%%%%%%%%%%%%%%%%%%%%%%%%%%%%%%%%%%%%%%%%%%%%%%%%%%%%%%%%%%%%%%%%%%%%%%%%%%%

%%%%%%%%%%%%%%%%%%%%%%%%%%%%%%%%%%%%%%%%%%%%%%%%%%%%%%%%%%%%%%%%%%%%%%%%%%%%%%%%

%%%%%%%%%%%%%%%%%%%%%%%%%%%%%%%%%%%%%%%%%%%%%%%%%%%%%%%%%%%%%%%%%%%%%%%%%%%%%%%%
% \bibliographystyle{IEEEtran}
% \bibliography{references}
%\balance

\end{document}